\documentclass[lettersize,journal]{IEEEtran}
\usepackage{amsmath,amsfonts}
\usepackage{algorithmic}
\usepackage{algorithm}
\usepackage{array}
\usepackage[caption=false,font=normalsize,labelfont=sf,textfont=sf]{subfig}
\usepackage{textcomp}
\usepackage{stfloats}
\usepackage{url}
\usepackage{verbatim}
\usepackage{graphicx}
\usepackage{xcolor}
\begin{document}

\title{Demystifying the Physics of Deep Reinforcement Learning-Based Autonomous Vehicle Decision-Making}

\author{Hanxi Wan,
        Pei Li, and
        Arpan Kusari,~\IEEEmembership{Member,~IEEE}
        % <-this % stops a space
\thanks{H.Wan is with the Department of Robotics, University of Michigan (e-mail:wanhanxi@umich.edu).}
\thanks{P. Li is with the Department of Civil \& Environmental Engineering, University of Wisconsin-Madison (e-mail:pei.li@wisc.edu).} 
\thanks{A. Kusari is with the University of Michigan Transportation Research Institute, University of Michigan (Corresponding author, e-mail:kusari@umich.edu). }% <-this % stops a space
\thanks{Manuscript received April 19, 2021; revised August 16, 2021.}}

\markboth{Journal of \LaTeX\ Class Files,~Vol.~14, No.~8, August~2021}%
{Shell \MakeLowercase{\textit{et al.}}: A Sample Article Using IEEEtran.cls for IEEE Journals}


\maketitle

\begin{abstract}
With the advent of universal function approximators in the domain of reinforcement learning, the number of practical applications leveraging deep reinforcement learning (DRL) has exploded. Decision-making in autonomous vehicles (AVs) has emerged as a chief application among them, taking the sensor data or the higher-order kinematic variables as the input and providing a discrete choice or continuous control output. There has been a continuous effort to understand the black-box nature of the DRL models, but so far, there hasn't been any discussion (to the best of authors' knowledge) about how the models learn the physical process. This presents an overwhelming limitation that restricts the real-world deployment of DRL in AVs. 
Therefore, in this research work, we try to decode the knowledge learnt by the attention-based DRL framework about the physical process. We use a continuous proximal policy optimization-based DRL algorithm as the baseline model and add a multi-head attention framework in an open-source AV simulation environment. We provide some analytical techniques for discussing the interpretability of the trained models in terms of explainability and causality for spatial and temporal correlations. We show that the weights in the first head encode the positions of the neighboring vehicles while the second head focuses on the leader vehicle exclusively. Also, the ego vehicle's action is causally dependent on the vehicles in the target lane spatially and temporally. Through these findings, we reliably show that these techniques can help practitioners decipher the results of the DRL algorithms.
\end{abstract}

\begin{IEEEkeywords}
Autonomous vehicles, deep reinforcement learning, attention, explainability, causality.
\end{IEEEkeywords}

\section{Introduction}
% AVs decision-making and RL
\IEEEPARstart{T}{he} objective of autonomous vehicles (AVs) is to safely navigate the uncertain dynamic environment using onboard sensors and some prior knowledge about the surroundings. To do so, the AVs utilize the paradigm of ``sense-plan-act" that has been central to robotics for the past half-a-century \cite{hartavi2021reliable} - 
\begin{itemize}
    \item sensing and perception module uses the high-dimensional sensing data to determine the static objects for localizing themselves and the dynamic objects and their trajectories for obstacle avoidance,
    \item decision-making module utilizes the localization information, kinematics of the dynamic objects, and the goal of the AV to take the optimal action, and
    \item control module executes the action in the form of path planning and path following.
\end{itemize}
Of these three high-level modules, the perception and control modules have reached a certain stage of high precision based on the deep learning architectures and the model-based methods respectively. However, in the realm of decision-making, the algorithms need to reason about the uncertainties present in the sensor data along with the incompleteness of the data (due to occlusion) among other factors, and provide an optimal action under these constraints. Thus, a stochastic cost function needs to be optimized for a high-dimensional, sequential process. Typically, a finite state machine (FSM) has been used in practical AV applications for decision-making. FSM uses discrete decision choices based on the current state and predicted future states of the ego and other vehicles based mostly on hand-tuned parameters \cite{zhang2017finite}. As can be easily seen, even though the solution is practical, it is neither scalable nor it can probabilistically handle the uncertainties. 

Reinforcement learning (RL) is a probabilistic formalism of sequential decision-making, where an agent learns the policy (mapping of states to actions) which maximizes the expected rewards in an environment \cite{sutton2018reinforcement}. For continuous state spaces and discrete/continuous action spaces, universal function approximators such as neural networks have been proposed with outstanding achievements - AlphaGO defeated the reigning GO champion \cite{silver2016mastering} and the Atari games were played at superhuman abilities by the deep reinforcement learning (DRL) agents \cite{mnih2013playing}. Researchers have reported great gains in AV decision-making using DRL as well, with the learned policy adapting to uncertainty in the future state space \cite{kiran2021deep}. 

So, a central question arises - \textit{if DRL has shown such exceptional performance in decision-making tasks including in AVs, why are state-based decision models such as FSM still being used primarily?} 

The answer lies in the black-box nature of the DRL and the difficulty that arises in understanding the intention (based on physics) behind the action. For a safety-critical application such as AVs, a sub-optimal but explainable solution is clearly preferred. In this paper, we hope to provide some scientific insights into that black box using some interpretation tools. Different from other papers in the explainability domain, we focus on deciphering the physics that the model learns, which is important for the downstream engineering tasks. Specifically, we concentrate on the attention-based mechanism which provides weights on the states (in this case, the neighboring vehicles) to interpret the policies. Our novel contributions are as follows:
\begin{itemize}
    \item we provide a benchmarking for the attention-based model with respect to a baseline proximal policy optimization (PPO) model in terms of kinematic variables, including average velocity, distance traveled, and time-to-collision (TTC). 
    \item we look at the explainability of the attention-based model in terms of spatial and temporal correlations and show that the specific discrete events (going straight to lane change decision) can be mapped by a change in attention weights. 
    \item we also look at the causality of the attention-based model using Iterative Causal Discovery (ICD) \cite{rohekar2022iterative} for the spatial changes and the temporal ICD \cite{rohekar2023temporal} for the temporal changes. We show that the particular lane change has a specific causal marker in the leader and follower in the target lane (i.e. left leader and left follower in the left lane). Also, temporal causality can be ascertained by the ego vehicle's influence on the previous time step along with the leading vehicle's influence. 
\end{itemize}

The paper is organized as follows: Section \ref{sec:background} provides the background in AV decision-making and where explainability and causality are applied in this process. Section \ref{sec:method} provides the state and action representation, the baseline and multi-head attention-based models used in this study, the reward function employed, the training process utilized, and the methods used in providing explainability and causality. Section \ref{sec:results} compares the results for the baseline and the attention-based models and showcases results in explainability and causality using spatial and temporal correlations. Finally, in Section \ref{sec:conclusion}, we provide some discussion, conclusions and future work. 

\section{Background}
\label{sec:background}
\subsection{AV Decision-making}
AV decision-making can be broken down into two different types - classical and learning-based methods~\cite{liu2021decision}. Classical methods, such as FSM~\cite{leonard2008perception,ziegler2014making}, Model Predictive Control (MPC)~\cite{wang2019path,cheng2019longitudinal,yu2021mpc}, and game theory~\cite{li2017game}, etc., have been widely used for AV decision-making tasks including path tracking, car-following, and lane-changing. The classical methods offer significant advantages in terms of simplicity and strong interpretability. Moreover, methods such as MPC and game theory can model the interactions between different traffic participants. However, classical methods may have poor performance in complex scenarios as their parameters are usually hand-tuned after simulation and field testing~\cite{kuutti2020survey}.

Learning-based methods, particularly deep learning, offer several advantages for vehicle control over classical methods. Deep learning methods excel in handling complex and non-linear control and can generalize previously learned rules to new scenarios~\cite{kuutti2020survey}. Two common deep learning methods, including supervised learning and reinforcement learning, are commonly used in existing studies. Supervised learning utilizes labeled data, where an expert demonstrates the task for training purposes~\cite{kuutti2020survey}. The labeled data contains observation-action pairs, such as images from onboard cameras and corresponding steering angle values. During training, the model learns how to act based on the observations and the actions. Supervised learning is easy and efficient to implement. However, it may not generalize well as it is heavily dependent on the training data. Differently, reinforcement learning allows the model to learn actions through trial and error and does not require labeled data. Therefore, models developed using reinforcement learning can generalize well to new scenarios. Among existing deep-learning methods, model-free DRL methods are becoming more and more popular as they allow adaptive and flexible learning from raw experiences without needing a predefined model, enhancing real-time performance in dynamic environments. They have been used in various AV control tasks, including lane-change~\cite{he2022robust,chen2019attention}, car-following~\cite{zhu2018human}, path-tracking, and platooning~\cite{prathiba2021hybrid}. DRL methods treat AVs as agents, which interact with the environments and learn the consequences of their actions through experience via algorithms such as policy gradient, Q-Learning, etc. As they are model-free, they do not require specific models to measure the environments, and the agents can learn directly from experience. This makes them well-suited for complicated driving environments where the dynamics are unknown or complicated to model.

\subsection{Explainability and Causality in AV Decision-making}

Although deep learning models have achieved promising accuracy in AV decision-making. There have been concerns over their transparency and accountability~\cite{omeiza2021explanations} since they are known as "black-box" models. Therefore, explainability and causality have become necessary criteria while evaluating deep learning-based AV decision-making systems. Specifically, explainability refers to the ability to explain and justify the underlying rationale of AVs' decisions. Differently, causality explores the cause-and-effect relationships between variables, such as the actions of neighbor vehicles and the actions of AVs. It aims to determine whether changes in one variable directly lead to changes in another variable, rather than just identifying similarities. Although DRL-based models have made significant progress and achieved superior performance in AV control compared to classical models, one major limitation of DRL-based models is the lack of explainability and causality. To solve this problem, explainable reinforcement learning, known as XRL, has been proposed and has gained a lot of attention in recent years. The main focus of XRL is on the reasoning behind the decisions or predictions made by the RL model~\cite{phillips2020four}. Specifically, the majority of XRL methods function by mimicking and simplifying a complex model using a more explainable model, such as tree-based models~\cite{puiutta2020explainable,schmidt2021can,cui2022interpretation}. Specifically, a DRL policy is first trained for controlling the AV. Then, a decision tree model is developed to explain the model using the observations of the agent as inputs and the decisions generated by the policy as outputs. This approach is easy to implement and interpret, however, the explainability may not reveal the actual knowledge learned by the agent. To further improve the explainability of DRL-based AV control models, additional research needs to be done on improving the inherent explainability of the DRL model rather than using a simple model to imitate it. Moreover, little research has been done to understand the causality behind deep learning-based methods for AV decision-making. \cite{kim2017interpretable} attempts to utilize a causality test for understanding the causality of a vehicle's actions using image data. A visual attention model is trained end-to-end from images to the steering angle. The causality is then decided by iteratively masking attention regions in images and evaluating the impact on the model's performance. This casual filtering technique produces a simpler and more accurate map of visual saliency by filtering out spurious attention regions. However, causality was not further analyzed but served as a technique to improve the interpretation of the attention maps.

\section{Methodology}
\label{sec:method}
From the background work, it is abundantly clear that while there have been efforts made to understand the DRL models used in AV decision-making, to the best of the authors' knowledge, there has been no concerted effort made to decode the state space and actions produced by DRL with respect to the kinematics. Similarly, there has been some work done in marrying the field of causality with RL, but it has not been applied to the spatio-temporal nature of AV decision-making. In this section, we present our method to train a reinforcement learning-based AV decision-making in a simulated highway environment task and examine the attention mechanism to infer interpretability. % Section \ref{subsec:state} shows the state and action representation and section \ref{subsec:model} gives a detailed view of our model architecture. Section \ref{subsec:reward} shows how we define the reward function and section \ref{subsec:train} describes our two-stage training method. 

\subsection{State and action representation}
\label{subsec:state}
We choose the observation space of the environment $o_t \in \mathbb{R}^{7\times5}$ as a vector representation of the scene. Referred to as affordance features in the literature \cite{chen2015deepdriving}, it provides information on the ego vehicle as well as the six neighboring vehicles (leading and following vehicles in the left/same/right lane of the ego vehicle). For each vehicle, there are 5 features: presence showing if the corresponding vehicle exists, longitudinal/lateral position, and longitudinal/lateral velocity. The position of the neighboring vehicle is represented as the distance to the ego vehicle, and all the distances and velocities are normalized. %Figure \ref{fig:obs_vis} visualizes the environment with the yellow standing for ego vehicle, and blue stands for neighboring vehicles. 
The action space of the environment $a_t \in \mathbb{R}^2$ is a two-dimensional vector that directly controls the throttle and steering angle of the vehicle. The vehicle dynamic is simulated using the Kinematic Bicycle Model (\cite{bicyclemodel}).
% \begin{figure}
%     \centering
%     \includegraphics[width=0.6\textwidth]{figures/Observation_example.png}
%     \caption{Visualization of the Highway-env simulation environment with the yellow rectangle representing the ego-vehicle and the blue rectangles representing the other vehicles.}
%     \label{fig:obs_vis}
% \end{figure}

% \begin{table}[ht]
% \centering
% \begin{tabular}{lllll}
% \hline
% Presence     & x    & y    & vx    & vy \\ \hline
% 1.  &  1.     & 0.2155  & 0.9526  & -0.30673212 \\
% 1.  &  0.0346 & 0.1845  & -0.3868 & 0.30673212 \\
% 1.  &  0.0612 & 0.0245  & -0.3591 & 0.30673212 \\
% 0.  &  0.     & 0.      & 0.      & 0.         \\
% 1.  & -0.3586 & 0.1690  & -0.4310 & 0.76405364 \\
% 1.  & -0.1716 & 0.0245  & -0.3815 & 0.30495825 \\
% 1.  & -0.0373 & -0.1355 & -0.3736 & 0.30673212 \\ \hline
% \end{tabular}
% \label{tab:obs}
% \caption{Observation Example}
% \end{table}

% \subsubsection{action space}
% The action space of the environment $a_t \in \mathbb{R}^2$ is a two-dimensional vector that directly controls the throttle and steering angle of the vehicle. The vehicle dynamic is simulated using the Kinematic Bicycle Model (\cite{bicyclemodel}).

\begin{figure*}
    \centering
    \includegraphics[width=\textwidth]{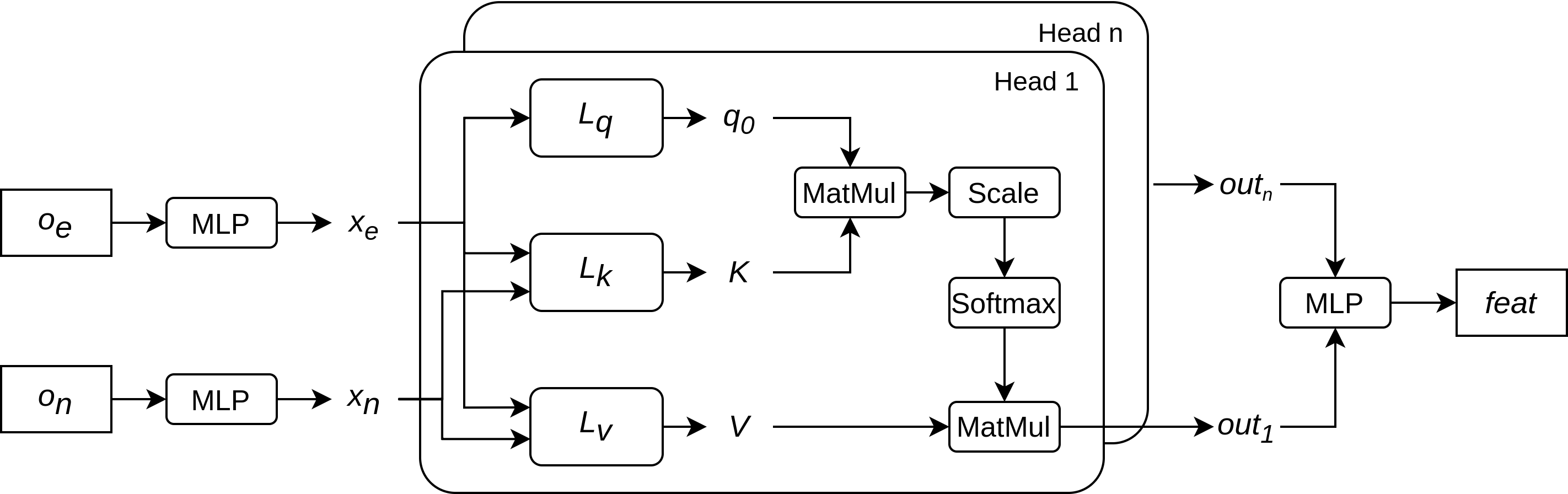}
    \caption{Architecture of the attention-based feature extractor.}
    \label{fig:model_arch}
\end{figure*}

\subsection{Model architectures}
The model is separated into two main parts: the feature extractor for extracting features from observations, and an action predictor for mapping features to actions. We propose two agents, a baseline agent and an attention-based agent. They have different feature extractor architectures, but they both use fully connected networks for the action predictor.

\subsubsection{Baseline Model}
The baseline model is a multi-layer perceptron (MLP) model that inputs the observation vector and outputs the action predictor. We use the baseline model as an expert for the attention-based agent. Therefore, the action predictor for the baseline agent is scaled so that the baseline agent and attention-based agent have a comparable number of trainable parameters.

\subsubsection{Multi-Head Attention-Based Model}
\label{sec:model}
To provide importance weights of the input features, we use a multi-head attention-based model inspired by \cite{leurent2019social} to extract features. However, different from the original idea, we vary the number of attention heads to find the optimal number of heads required for optimal use. Figure \ref{fig:model_arch} shows an overview of the attention-based model.

Firstly the observation goes through an ego-attention feature extractor. The ego vector $o_e \in \mathbb{R}^{1\times5}$ and neighboring vectors $o_n \in \mathbb{R}^{6\times5}$ are encoded by an MLP layer to get the embeddings of size $d_k$. Then the embeddings are fed into a multi-head ego-attention layer to extract features of the observation. Different from the self-attention mechanism which produces output for each input element, the ego-attention layer only produces one single output corresponding to the ego vehicle. Each vehicle's embeddings go through a linear layer $L_k \in \mathbb{R}^{d_x \times d_k}$ to get its key $k_i$, and a linear layer $L_v \in \mathbb{R}^{d_x \times d_v}$ to get its value $v_i$. The ego embedding alone goes through a query linear layer $L_q \in \mathbb{R}^{d_x \times d_k}$ to get its query $q_0 \in \mathbb{R}^{1 \times d_k}$. The attention matrix $attn \in \mathbb{R}^{1\times 7}$ is the modified similarities between the query from ego vehicle $q_0$ and the keys from neighboring vehicles $k_i$, shown in Equation \ref{eq:attn}. The inner product of $Q=q_0$ and $K_T = [k_0, k_1, ..., k_6]^T$ is the similarity, which is then scaled by the square-root-dimension and normalized with a softmax function $\sigma$. After that, the attention matrix is multiplied by the value $V = [v_0, v_1, .., v_6]$ to get the final output for one head. The outputs of all head $out_i \in \mathbb{R}^{1\times d_v}$ are fed through a linear layer to form the combined feature representation from different heads $feat \in \mathbb{R}^{1\times 64}$.

\begin{equation}
    \label{eq:attn}
    attn = \sigma (\frac{QK^T}{\sqrt{d_k}})
\end{equation}

% The baseline agent has a similar architecture as the attention-based agent except that its feature extractor is replaced by a MLP. It has a similar number of parameters (42629) as the attention-based agent (42309).

% \subsubsection{Action Predictor}

% \textcolor{red}{TODO: PPO model - attention mechanism - multi head
%                 }

\subsection{Reward Function}
\label{sec:reward}
For any RL algorithm, the reward function (referred to as the objective function or the utility function in other optimization literature) is the most crucial step - the agent optimizes its expected sum of rewards based on the reward function. However, determining the reward function is not straightforward in the best of situations: in many real-world applications, rewards are delayed or sparse. In the case of AVs, a certain acceleration or steering angle can lead to a crash in the large time horizon which makes it difficult to assign credit/blame on a particular action choice. Therefore, in RL literature, several studies have used reward shaping \cite{ng1999policy} to provide additional expert knowledge, to improve learning speed and convergence.

Inspired by the reward function in \cite{baheri2020deep}, we design our reward function as a vector, as shown in Eq. \ref{eq:reward}. This reward function utilizes multiobjective RL (MORL) where tradeoffs between conflicting objective functions. Our reward function takes into consideration the velocity, the lateral position in the lane, the action, and the distance from the leading vehicle by creating a scalarization function \cite{kusari2020predicting}:

\begin{equation}
\label{eq:reward}
r= \begin{cases}0, \quad\quad \text { if TTC }< 0.2 \text { or } v_x<20 \\ w_v \cdot r_v + w_y \cdot r_y + w_a \cdot r_a + w_x \cdot r_x, & \text { otherwise }\end{cases}
\end{equation}

\begin{equation}
    \begin{aligned}
    r_v &= e^{-\frac{((v_y - v_{des})/10)^2}{10}}, 
    r_y = e^{-\frac{(y-y_{des})^2}{30}},
    r_a = e^{-(3*a_{steering})^2}, \\
    r_x &= e^{-\frac{(d_{lead} - d_{safe})^2}{10d_{safe}}} \text{,  if  } d_{lead} < d_{safe},
    \end{aligned}
\end{equation}
where $v_{des}$ is the desired speed, $y_{des}$ is the center of the desired lane determined by the previous series of actions and the current lane, $d_{lead}$ is the distance to leading vehicle, $d_{safe}$ is the safe longitudinal distance, and $w_v, w_y, w_a, w_x$ are the weights for different rewards. We performed a set of controlled experiments to determine the weights of the various factors and chose a set of weights that balanced the driving behavior with the safety of the vehicle. 

\subsection{Training Process}
\label{subsec:train}

\subsubsection{Reinforcement learning}
%Reinforcement learning is a general probabilistic framework for sequential decision-making under uncertainty. 
Based on the feature we extracted, we train an action predictor in reinforcement learning using the proximal policy optimization (PPO) algorithm \cite{ppo}. We train our agent in highway-env using PPO as the reinforcement learning algorithm that takes the $feat$ from the feature extractor as input and outputs an action $a \in \mathbb{R}^{2}$. PPO is an RL architecture that improves training stability by avoiding large policy updates. The intuition is that performing smaller policy updates during training converges to an optimal solution more often. Therefore, the change is clipped to $\epsilon$ over the current policy. Its loss function is

\begin{equation}
\begin{aligned}
    &L_t^{CLIP+VF+S}(\theta) = \hat{\mathbb{E}_t} [ L_t^{CLIP}(\theta) - c_1 L_t^{VF}(\theta) + c_2 S[\pi_\theta] (s_t)],\\
    &L_t^{CLIP}(\theta) = \hat{\mathbb{E}_t} [\min ( r_t(\theta)\hat{A_t}, clip(r_t(\theta), 1 - \varepsilon, 1 + \varepsilon) \hat{A_t}],
\end{aligned}
\end{equation}
where $\hat{\mathbb{E}_t}$ is the expectation operator,
$L_t^{CLIP}$ is the clipped surrogate objective,
$L_t^{VF}$ is the squared error loss of the value function $(V_\theta (S_t) - V_t^{targ})^2$,
$S[\pi_\theta] (s_t)$ is an entropy bonus to ensure sufficient exploitation.
In $L_t^{CLIP}$, $\varepsilon$ is a hyperparameter,
$r_t(\theta) = \pi_\theta (a_t | s_t) / \pi_{\theta old}(a_t | s_t)$ is the probability ratio. The clipped loss function effectively avoids taking a large policy update.

\subsubsection{Imitation Pre-Training for Attention-based framework}
To decrease the overall training time and increase the sample efficiency of the attention-based algorithm, we devise a two-stage process to train our model. In the pre-training stage, imitation learning enables the model to learn policy efficiently based on the sampled trajectories from the baseline agent. In imitation learning, the agent learns from the trajectories and the corresponding actions provided by an expert. The primary objective of the pre-training phase is to accelerate the process of policy learning by imitating the behavior of the baseline agent.

During the pre-training phase, we generate a set of observation-action pairs using a baseline model. Then we train both the feature extractor and action predictor as a whole network by minimizing the minimum squared error loss of action and value between the model's prediction and the baseline model's reference. However, since imitation uses a sampling of observation and action pairs, the trained agent might get exposed to states not encountered in training. We then utilize the second reinforcement learning stage that enables the model to adapt to the stochastic nature of the environment as well as optimize its policy.

\subsection{Explainability using the attention mechanism}

Explainability in machine learning models aims towards ``making sense" to a human being at an acceptable level. Attention mechanisms have been used successfully towards providing explainability of deep learning models \cite{sundararajan2017axiomatic,smilkov2017smoothgrad, kobayashi2020attention}. 

They do so by generating an attention matrix, which essentially encodes the importance correlation - the magnitude of the explanation weights that precisely reflect the importance of the input data. High attention weights indicate that the corresponding features are important in the current observation and decisive in generating the policy. This, therefore, allows us to visualize and interpret how the model focuses on different parts of the input data and assigns varying levels of importance to different features.  

We would like to mention that while some researchers provide the notion that attention does not automatically encode explainability \cite{jain2019attention}, other researchers have countered their claims based on the meaning of ``explanation" which needs to take into account all the elements of the model \cite{wiegreffe2019attention}. % The attention weights highlight which features are useful in the decision-making process and thus provide an explanation of the model's prediction.

In our case, similar to \cite{leurent2019social}, we learn the attention matrices corresponding to the heads of the model. Essentially, for a state matrix comprised of the kinematics of the vehicles around a given ego vehicle, the DRL model should pay attention to the closest vehicles and also be attentive to the vehicle in direct conflict with the ego vehicle's future trajectory. Specifically, we look at the correlations between distances and weights across the heads to look at the spatial explainability. We also plot the weights over the entire run to look at the temporal explainability of the DRL agent based on the attention weights.

\subsection{Causality using Iterative Causal Discovery}
\label{subsec:icd_method}
While explainability based on the attention mechanism has been touted as sufficient, it doesn't suffice in discovering the reason why the model produces certain outputs given the observation. Also, the attention-based models suffer from a lack of explanation faithfulness \cite{jacovi2020towards} where there is a polarity inconsistency \cite{liu2022rethinking}, enhancing or suppressing the input. We, therefore, look towards applying causality in inferring causal relations using the attention matrix through the Iterative Causal Discovery (ICD) \cite{rohekar2022iterative}. In ICD, session-specific causal graphs are learned from attention to the possible presence of latent confounders. The authors utilize the graphs to learn neural recommenders. 

Causal discovery uses certain assumptions - chief among them are the causal Markov and faithfulness, with no parametric assumptions placed on the distribution of the data. Given these assumptions, tests of conditional independence (CI-tests) can be performed to learn the causal structure. The CI-test checks whether the two nodes are independent given a conditioning set of nodes using partial-correlation CI testing after evaluating the correlation matrix from the attention matrix $A$. The covariance matrix is defined as $K = AA^T$ and the correlation coefficients are defined as $\rho_{i, j} = K_{i,j}/\sqrt{K_{i,i}.K_{j,j}}$.

Based on running the CI-tests, a partial ancestral graph (PAG) can be constructed, which defines a set of causal graphs where the causal relations definitely exist. In a PAG, there are three types of edge-marks at a given node X - an arrow-head ’$\rightarrow$X ’, a tail ’—–X ’, and a circle ‘—$\circ$ X’ which represents an edge-mark that cannot be determined given the data. 

We look at ICD to provide causal relations between the vehicles in the event of a lane change. Therefore, we separate the lane changes based not only on left and right lane changes but also, on the origin lane. We aggregate the attention matrices from the two heads for each type of lane change and the source lane for 10 seconds before the lane change. We then conduct the CI-tests and construct PAG for each lane change subtype.

Similar to the previous subsection, we look at the temporal nature of the causal relationships for the neighboring vehicles using the temporal ICD \cite{rohekar2023temporal}.

\section{Results}
\label{sec:results}
\begin{figure}
    \centering
    \includegraphics[width=0.4\textwidth]{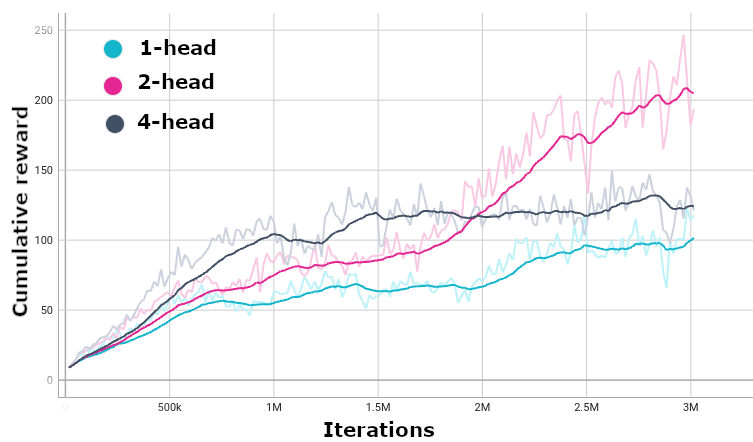}
    \caption{Comparing rewards over iterations for different heads}
    \label{fig:reward}
\end{figure}
% \begin{wrapfigure}{r}{0.5\textwidth}
\begin{figure}
    \centering
    \includegraphics[width=0.5\textwidth]{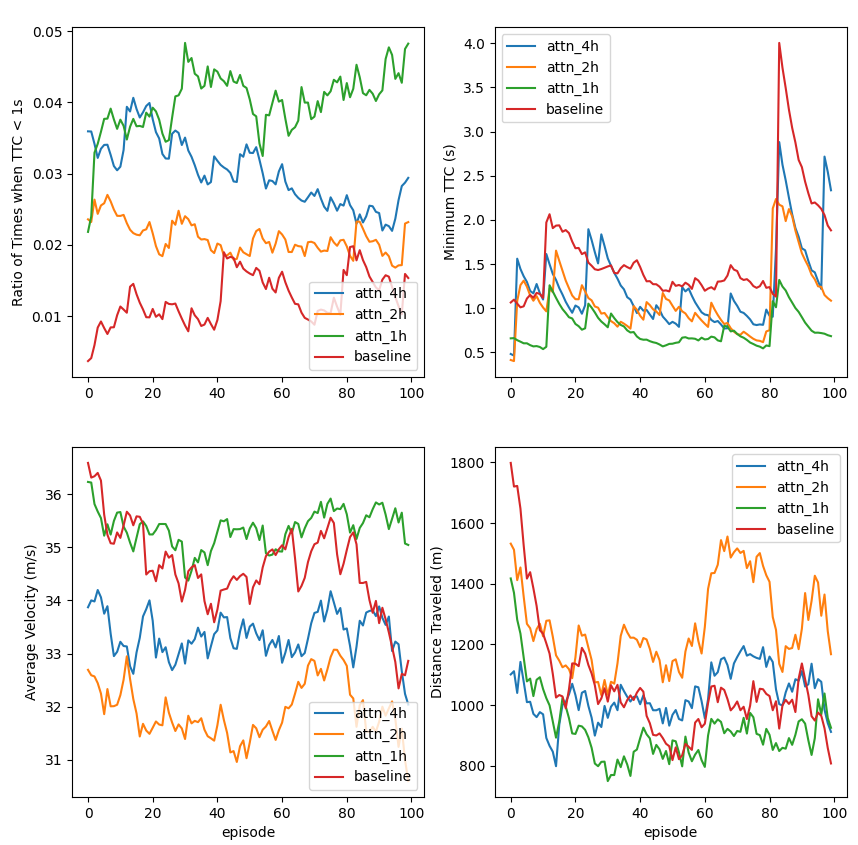}
    \caption{Baseline model vs the attention-based model using common driving metrics}
    \label{fig:benchmark}
\end{figure}
In this section, we aim to provide interpretations of the actions of the DRL given the state. To do this, we use an open-source AV simulation environment, highway-env \cite{highway-env}, developed for highway driving tasks. The agent is placed on a straight three-lane highway with a user-defined number of other vehicles spawned randomly in every episode. The other vehicles have IDM (Intelligent Driver Model)-based longitudinal control \cite{treiber2000congested} with a discrete lane change decision based on the MOBIL model (Minimizing Overall Braking Induced by Lane change) \cite{kesting2007general}. This environment provides the agent with a kinematic representation of ego and neighboring vehicles, and the agent can take either discrete lane change decisions or continuous acceleration and steering angle information at each step.

We provide results first by benchmarking the multi-head attention-based DRL against the baseline model setup in terms of common driving metrics. One of the first questions we aim to answer is the number of heads that can provide the optimal output. In order to answer this question, we attempt to vary the number of heads to be between 1, 2, and 4. We also interpret the results of the model in two adjoining functionalities: explainability and causality; using both spatial and temporal correlations.

\subsection{Benchmarking DRL agent performance}
We train the baseline model and the 1-, 2- and 4-head attention models over the 3M global steps (3-head did not provide any results). Figure \ref{fig:reward} shows the reward over the iterations for the different heads. We find that the 2-head fares better compared to others in collecting rewards over the course of the training. 

We also compare the performance of the trained baseline and the multi-head attention-based models on a few common driving metrics- 
\begin{itemize}
    \item number of times in an episode that the TTC is below 1 sec as a ratio of the total episode length,
    \item minimum TTC in seconds during an episode,
    \item average velocity in an episode in meter/sec and,
    \item total distance traveled over an episode in m.
\end{itemize}
We run the trained models for 100 episodes in highway-env. As mentioned in section \ref{sec:method}, we use the baseline model as the expert and then train attention-based models additionally.
From the upper left subfigure of Figure \ref{fig:benchmark}, we find that the attention-based models generally get into more unsafe situations (given by the number of times the TTC goes below 1 sec) than the baseline model. The 1-sec threshold is based on the fact that for a vehicle at $35 m/s$ speed, reaction time of 100 ms, and deceleration of $10~m/s^2$, it takes the vehicle about 3 seconds to stop completely.
The upper right subfigure of Figure \ref{fig:benchmark} provides the minimum TTC in the episode which is highest for the baseline model. The low TTC indicates that the attention-based models are more likely to demonstrate risk-taking behaviors.
For the average velocity given in the lower left subfigure, the baseline model fares slightly better than the 2- and 4-head attention-based models while 1-head is the fastest consistently.
The lower right subfigure indicates the distance traveled over an episode without having a crash. Interestingly, the 2-head model performs the best consistently over the baseline and other attention-based models. Overall, our experiments show that adding attention weights skews the model towards being riskier drivers, even though its not linear. The 2-head model, for example, is more conservative than the 1- and 4-head models in terms of TTC and average speed. Therefore, we choose the 2-head model for further analysis.  

% \begin{wrapfigure}{r}{8cm}
\begin{figure}
    \centering
    \includegraphics[scale=0.35]{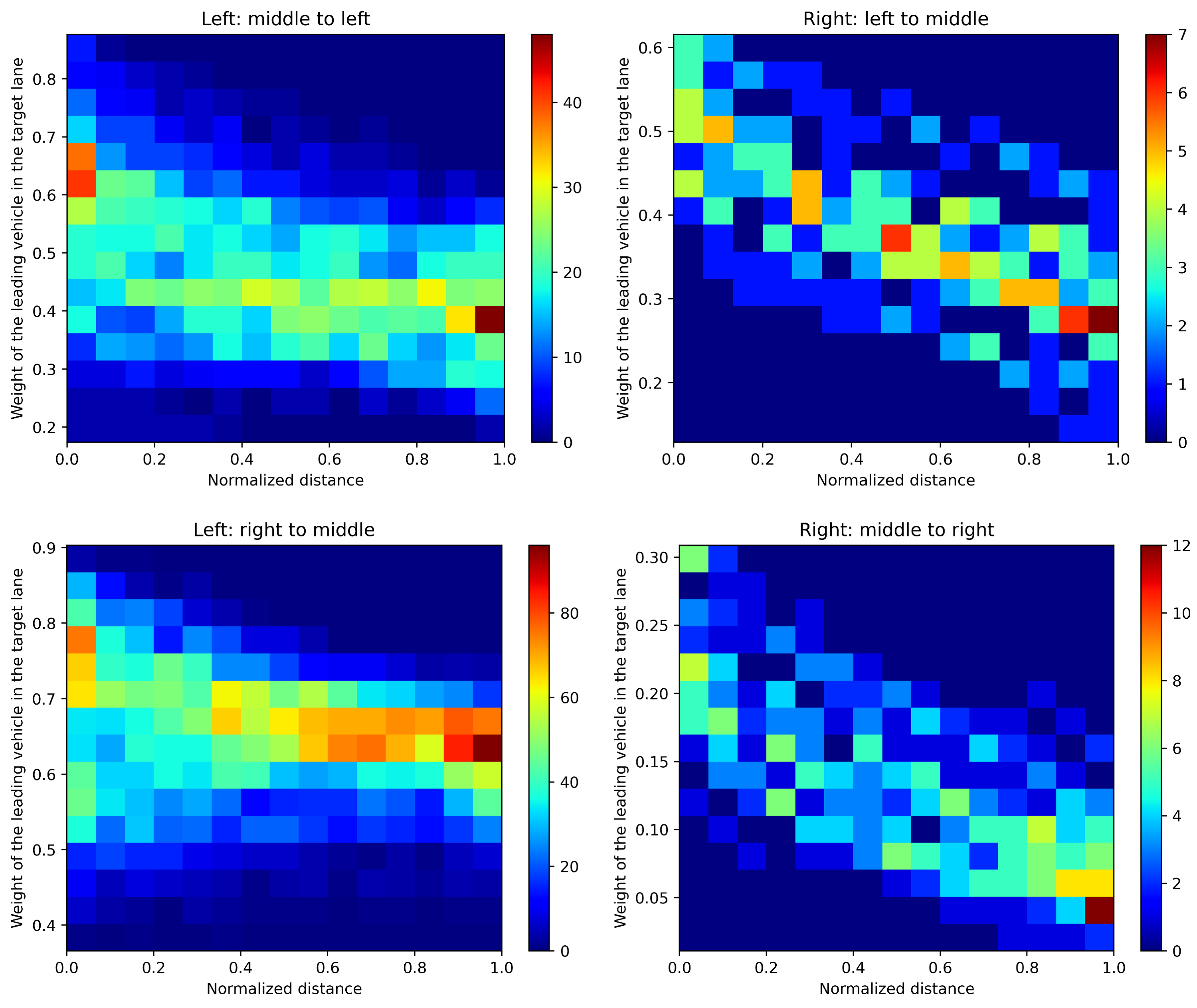}
    \caption{2D histogram showing the change in attention weights with respect to normalized distance to the leader vehicle during cumulative left and right lane changes - the weights can be shown to have direct correlation with distance although the rate of change differs significantly}
    \label{fig:attention_weight_lanechange}
% \end{wrapfigure}
\end{figure}

% \begin{table}[ht]
% \centering
% \begin{tabular}{lllll}
% \hline
% Agent     & Crash Rate    & Distance Travelled\\ \hline
% RL        & ? & ? \\
% baseline  & ? & ? \\ \hline
% \end{tabular}
% \caption{Performance Comparison.}
% \label{tab:perf}
% \end{table}

\subsection{Explainability Using Attention Interpretation}
\label{subsec:attn_interpretaion}
We interpret the attention in two different ways. Firstly, we analyze the relation between the distance to the leading vehicle and the corresponding attention weights. Then, we analyze the attention pattern in a specific lane change scenario. 

\begin{figure}
    \centering
    \includegraphics[width=0.5\textwidth]{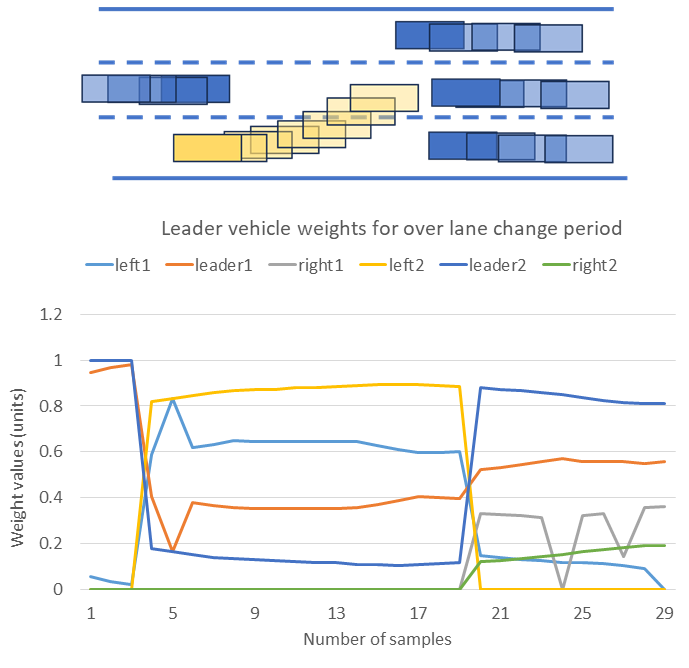}
    \caption{The top figure shows the lane change of the ego vehicle shown in yellow with the neighboring vehicles with the evolution into the future. The bottom plot shows the attention weights of the two heads for the lead vehicles in the three lanes mentioned in the legend.}
    \label{fig:temporal_explainability}
\end{figure}

\subsubsection{Spatial explainability of AV DRL}

We plot the attention weight of the leading vehicle in the target lane against the normalized distance from the AV to the leading vehicle for 10 seconds leading up to the lane change point. Both left lane changes and right lane changes are analyzed, and they are further divided based on the initial lane position of the AV. For example, two types of left lane changes are generated from the results, with the AV in the middle and right lane, respectively.

The relationships between the attention weights and the normalized distance are visualized using 2D histograms as shown in Figure~\ref{fig:attention_weight_lanechange}. We can see that the agent pays more attention to the leading vehicle when it's closer, which is reasonable since the closer distance indicates a higher risk of collision, thus this feature should be taken seriously. Moreover, taking the upper left figure as an example, the variations of the weights get smaller when the distance gets longer. This corresponds to the reality as the distance is related to the interaction between the leading vehicle and the AV, the interaction would be stronger for a shorter distance, which would result in a larger variation of attention weights. Lastly, the attention weight has a relatively left-skewed distribution when the distance is small, suggesting that the agent pays more attention when it is close to the leading vehicle. 

\subsubsection{Temporal explainability of AV DRL}
We investigate the temporal explainability by plotting the changes in attention weights when the AV performs a left lane change from the right lane to the middle lane as shown in Figure~\ref{fig:temporal_explainability}\footnote{The full video including the weights are given in \url{https://drive.google.com/file/d/1viFwTF_zxDWyRKnwOkUjKvtDHeey7gXz/view?usp=share_link}}. The weights of the left leader from the two attention heads gradually increase as the AV starts the preparation of the lane change. They then reach the maximum values when the AV performs the lane change. This result corresponds to the reality as the AV would pay more attention to the left leader since it needs to find the appropriate gap to change the lane. On the contrary, the weights of the leader gradually decrease as the AV starts performing the lane change. The AV pays more attention to the leader in the beginning as it needs to maintain a safe distance to the leader. However, it focuses less on the leader while performing the lane change as this action does not affect the leader directly.

\begin{figure*}
    \centering
    \includegraphics[width=0.8\textwidth]{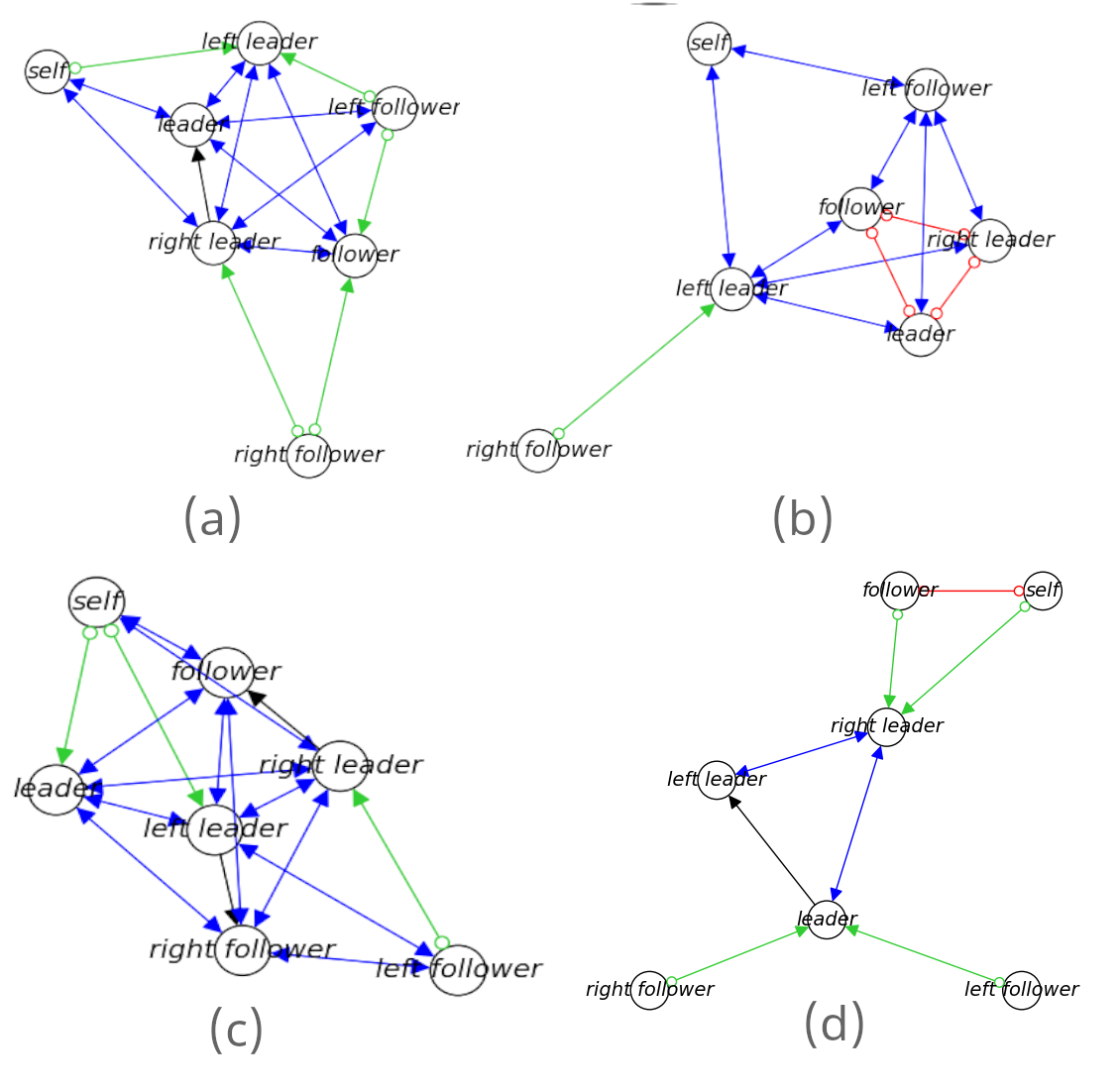}
    \caption{ICD graphs of lane changes divided by the source and target lane - (a) left lane change from right to middle lane, (b) left lane change from middle to left lane, (c) right lane change from left to middle lane, and (d) right lane change from middle to right lane}
    \label{fig:icd_graph}
\end{figure*}

\subsection{Causality}
In this section, we perform a causal analysis of the attention weights using ICD graphs \cite{icd-github}. Similar to the previous results, we collect the duration of the lane changes and classify them based on the type of lane change and the source lane.  

\subsubsection{Spatial causality}
To determine the spatial causality, we attempt to understand whether there is a causal relation between the ego-vehicle and certain neighboring vehicles during the lane change. We choose the first head since it focuses on all the neighboring vehicles instead of the second which focuses on the leader. As mentioned in Section \ref{subsec:icd_method}, there are three different types of edge-marks depending on the type of causal relationship: arrow-head for a causal relationship to the node, tail for causal relation from the node, and circle mark for no causal relation. Based on these edge-marks, we can get three edge colors - blue if both nodes affect each other, green if one node affects another, and red if there is not enough data to determine the relation. 

Figure \ref{fig:icd_graph} shows the results of the first attention head during the lane changes to understand which vehicle movement has a causal effect on the others. In the left lane changes (shown in the top row of the figure), the ego vehicle is causally related to the left leader and left follower, which influences how the ego vehicle changes lanes. Similarly, for right lane changes (bottom row), the ego vehicle is primarily influenced by the right leader. While these plots confirm our intuition that causal relationships can be formulated among the vehicles, we also find spurious correlations between other vehicles since lane change is inherently a spatiotemporal process. Therefore, we look at temporal measures of causality to infer a more meaningful relationship. 

\subsubsection{Temporal causality}
\begin{figure*}[ht]
    \centering
    \includegraphics[width=0.8\textwidth]{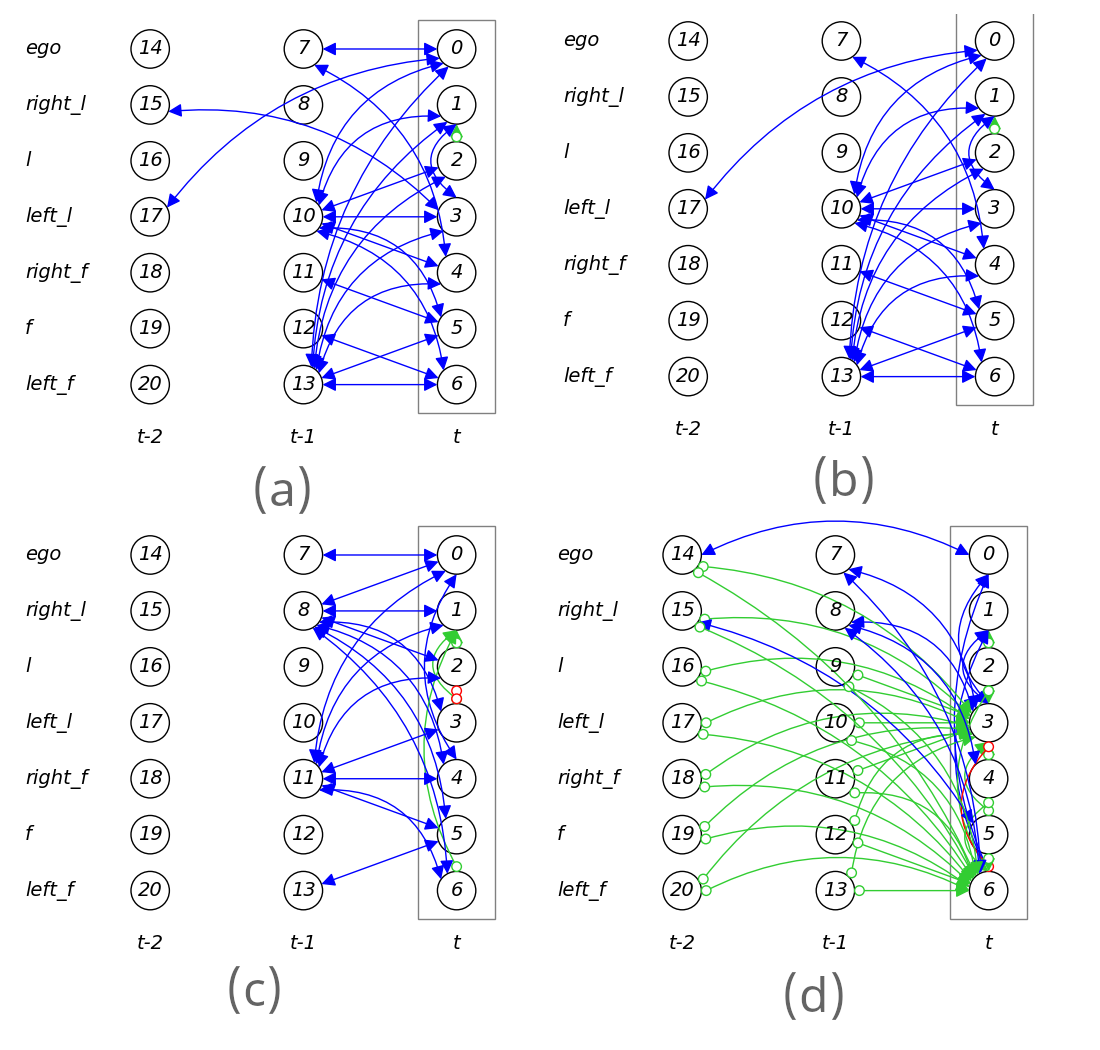}
    \caption{Time-series ICD graph of the lane changes with t-1 step signifying lane change point; t-2 step is 5 sec before and t is 5 sec after with - a) left lane change from right to middle lane, (b) left lane change from middle to left lane, (c) right lane change from left to middle lane, and (d) right lane change from middle to right lane}
    \label{fig:ts_icd}
\end{figure*}

To determine the temporal causal structures present in our data, we utilize the temporal ICD \cite{rohekar2023temporal}. We divide the time series into three discrete steps: lane change point is given as t-1, while t is 5 sec after the lane change point and t-2 is defined as 5 sec before the lane change point. From Figure \ref{fig:ts_icd}, we can see that the ego vehicle is primarily influenced by the leaders and followers in the target lane at t-1 step. Specifically, in the upper row of the figure, the ego vehicle is influenced by the left leader and left follower in the t-1 step while in the bottom row, the ego vehicle is influenced by the right leader and right follower in the t-1 step. This result clearly shows that a temporal causal relationship can be firmly established between the leader and the follower in the target lane. 

% t-2 is 3 timestep before the turning starts,
% t-1 is the time it starts to turn,
% t is 5 timestep after t-1. (need to be double-checked)

% \subsubsection{Evolution of ICD Graph during training}
% \textcolor{red}{TODO: include the changes in the graph during training. The connections in the graph were transitioning from a random pattern to the current pattern.}

\section{Discussion and conclusions}
\label{sec:conclusion}
In this paper, we examine the attention-based mechanism to provide answers regarding the decision-making used in DRL. While the attention-based DRL model does not substantially increase the performance of the model, it provides substantial benefits in interpreting the decision-making process. We show that explainability can be provided through the weights of the attention both in the spatial and the temporal domains. We also look at the causality of the neighboring vehicles - which shows that the particular lane change is triggered by the lead and the following vehicles. Our goal through this paper is to provide researchers with the analysis to provide some scientific interpretation of the actions provided by DRL and make the black-box neural network interpretable. Our future work would be in using the physically informed driving models to provide constraints on the reward function. 

\bibliographystyle{IEEEtran}
\bibliography{physics-explainability}

\newpage

% \vspace*{-4em}
\begin{IEEEbiography}[{\includegraphics[width=1in,height=1.25in,clip,keepaspectratio]{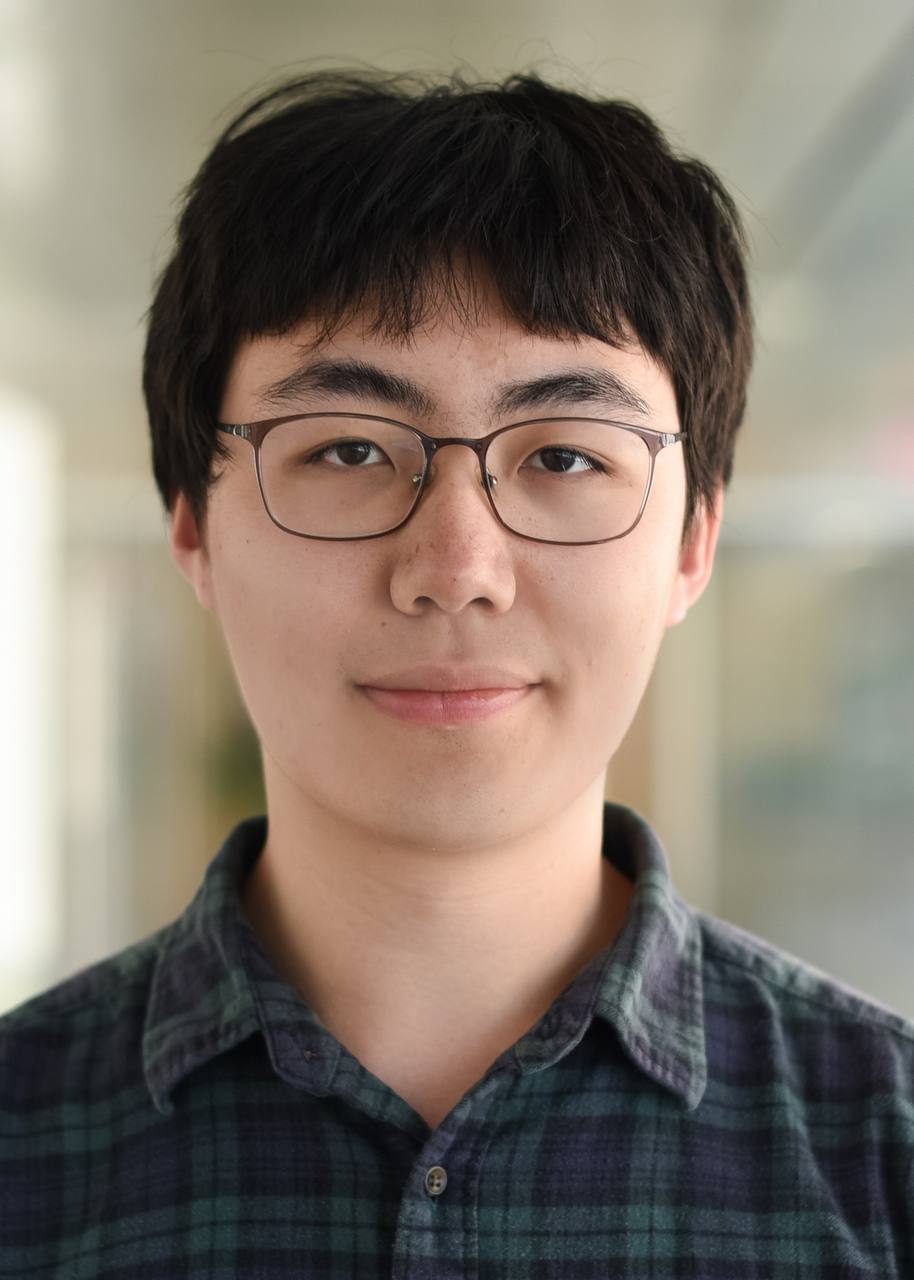}}]{Hanxi Wan} is a Master's student at the University of Michigan. He received his Bachelor's degree in Computer Engineering from the University of Michigan and Bachelor's degree in Electrical and Computer Engineering from Shanghai Jiao Tong University. His research interests include reinforcement learning, robot planning, and robot perception.
\end{IEEEbiography}
\begin{IEEEbiography}[{\includegraphics[width=1in,height=1.25in,clip,keepaspectratio]{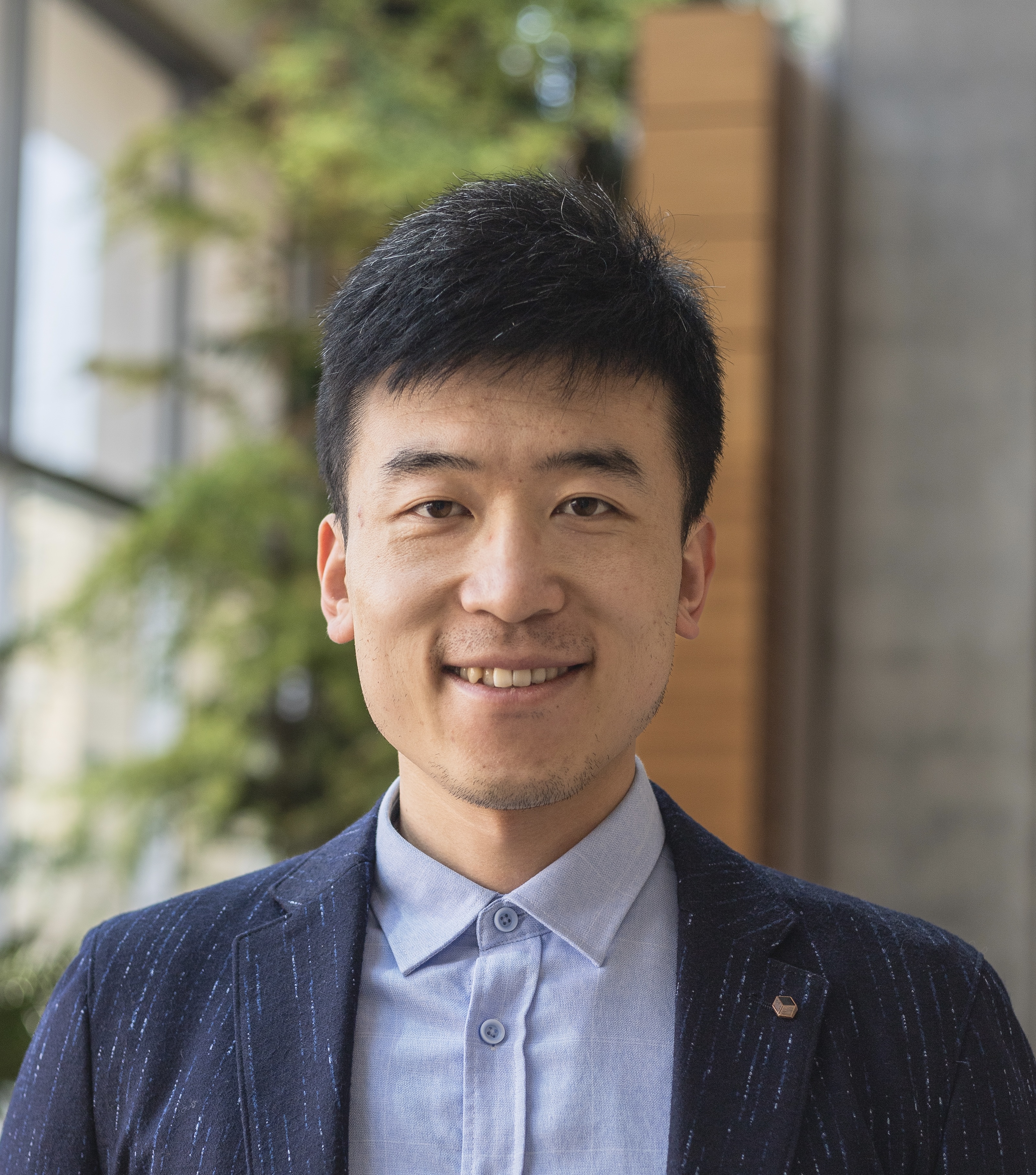}}]{Pei Li} is a Scientist at the University of Wisconsin-Madison. He received his Ph.D. in Civil Engineering from the University of Central Florida, Orlando, FL, USA. His research interests include traffic safety, deep learning, connected vehicles, and traffic simulation.
\end{IEEEbiography}
\begin{IEEEbiography}[{\includegraphics[width=1in,height=1.25in,clip,keepaspectratio]{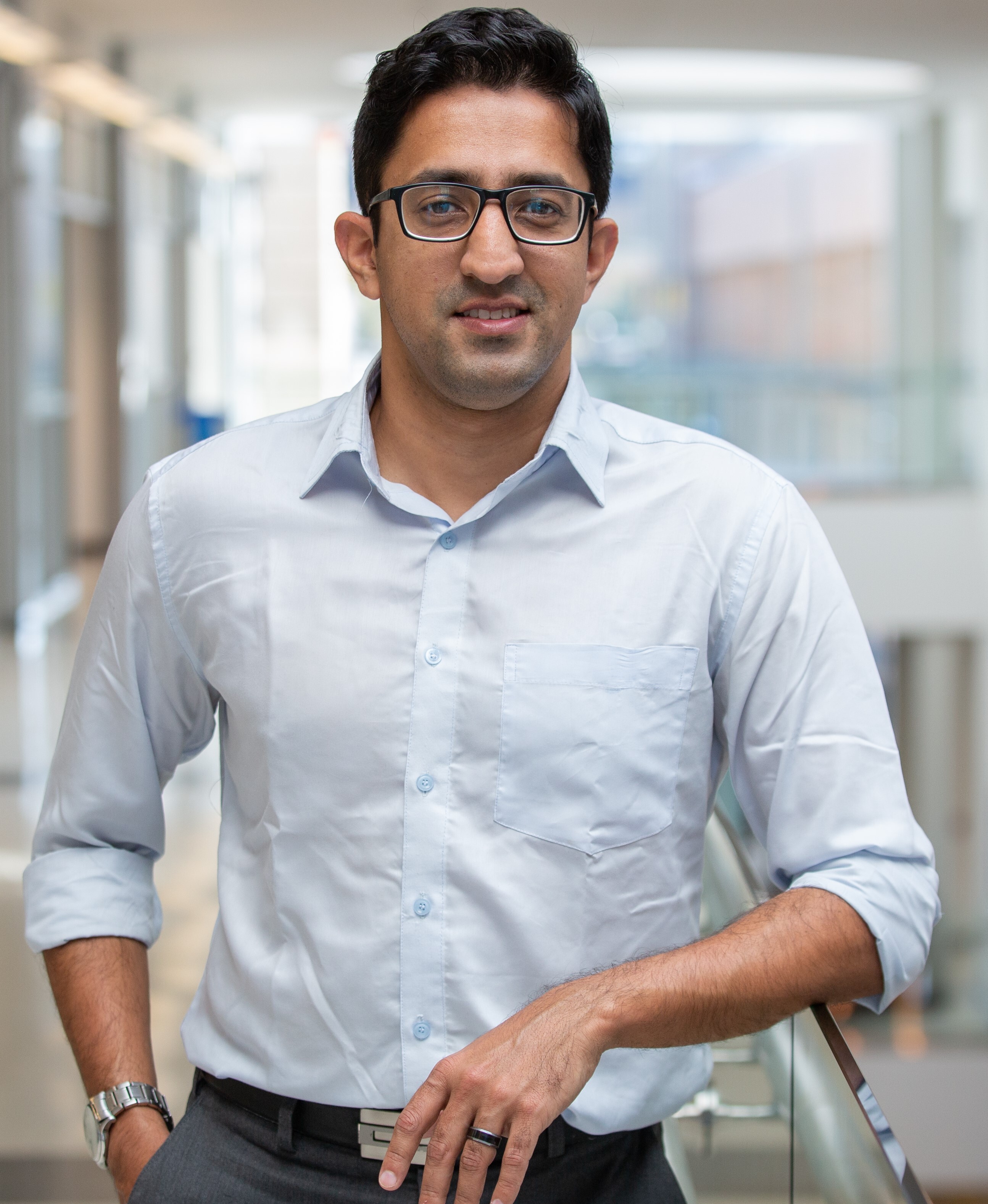}}]{Arpan Kusari} is an Assistant Research Scientist in the Engineering Systems Group at University of Michigan Transportation Research Institute. His primary interest is in Advanced Driver Assistance Systems (ADAS) and Automated Driving Systems (ADS) where he works on enhancing robustness of the software pipeline. He has been part of a number of federal and industry based projects in researching safety, sensor calibration and testing individual ADS systems such as localization. 
Previously, he spent five years at Ford Motor Company working collaboratively with researchers from MIT and University of Michigan; formulating research problems in autonomous vehicles. His doctoral research was in LiDAR mapping in the areas of sensor calibration, precise estimation of earthquake displacement and uncertainty quantification in the point cloud.
Dr. Kusari is a member of technical committees in SAE and ISPRS. 

\end{IEEEbiography}
 

% \end{IEEEbiography}



\vfill

\end{document}